%% file: main.tex
\documentclass[11pt]{article}

\usepackage[final]{acl}

\usepackage{times}
\usepackage{latexsym}
\usepackage{algorithm}
\usepackage{algpseudocode}
\PassOptionsToPackage{table}{xcolor}  
\usepackage{xcolor}
\usepackage{colortbl}
\usepackage{multirow}
\usepackage{makecell}
\usepackage{booktabs}

\usepackage[most]{tcolorbox}
\usepackage{fvextra}
\usepackage{fancyvrb}
\usepackage{cleveref}

\definecolor{sciorchhl}{RGB}{198, 224, 250}

\usepackage[T1]{fontenc}

\usepackage[utf8]{inputenc}

\usepackage{microtype}

\usepackage{graphicx}

\usepackage{amsmath}
\usepackage{amssymb}

\title{SciOrch: Learning to Orchestrate Expert LLMs for Solving Frontier Multimodal Scientific Reasoning Tasks}

\author{
  \bfseries Jingru Guo$^{1}$, Xiangyuan Xue$^{2}$, Lian Zhang$^{3}$, Wanghan Xu$^{4,5}$, Siki Chen$^{6}$, \\
  \bfseries Philip Torr$^{6}$, Wanli Ouyang$^{2,5,7}$, Lei Bai$^{5\dagger}$, Zhenfei Yin$^{6\dagger}$ \\[0.3em]
  $^1$Imperial College London \quad $^2$The Chinese University of Hong Kong \\
  $^3$University of Illinois Urbana-Champaign \quad $^4$Shanghai Jiao Tong University \\
  $^5$Shanghai AI Laboratory \quad $^6$University of Oxford \quad $^7$Shenzhen Loop Area Institute 
}

\begin{document}
\maketitle

{\renewcommand{\thefootnote}{\fnsymbol{footnote}}\footnotetext[2]{Corresponding authors.}}

\input{contents/00_abstract}

\input{contents/01_introduction}
\input{contents/02_related_work}
\input{contents/03_method}
\input{contents/04_experiments}
\input{contents/05_conclusion}

\input{contents/06_limitations}

\input{contents/07_ethical_considerations}

\bibliography{references}

\appendix
\input{contents/08_appendix}

\end{document}

%% file: contents/00_abstract.tex
\begin{abstract}
Frontier scientific reasoning remains a major challenge for large language models (LLMs), where even the strongest commercial systems fall short of expert-level performance. A closer look at model behavior reveals substantial complementarity that single-model evaluation hides: different frontier models excel on different question types, and no single model captures the full picture. We present \textbf{SciOrch}, a framework that trains a lightweight 8B model to orchestrate frontier LLMs for scientific reasoning. The orchestrator decomposes each question, delegates sub-problems to selected commercial models through API calls, and synthesizes a final answer. Training such an orchestrator is fundamentally harder than conventional agentic RL: each action triggers an API call that is expensive in both dollar cost and latency, making standard online rollouts infeasible. We address this with an MCTS-based approach, producing diverse orchestration trajectories, extracting per-node single-turn samples, and optimizing the orchestrator with GRPO-style training via REINFORCE++. On a 240-question test set spanning SGI-Reasoning and Scientists' First Exam, SciOrch reaches 56.66\% average accuracy, outperforming the strongest single commercial model by 3.74\% and the strongest multi-agent baseline by 3.33\%. It also attains the best accuracy on both SGI and SFE at a lower API cost than every self-consistency and debate baseline. Our code is available at: \url{https://github.com/llexieguo/SciOrch}.
\end{abstract}

%% file: contents/01_introduction.tex
\section{Introduction}

\begin{figure}[ht]
    \centering
    \includegraphics[width=1\linewidth]{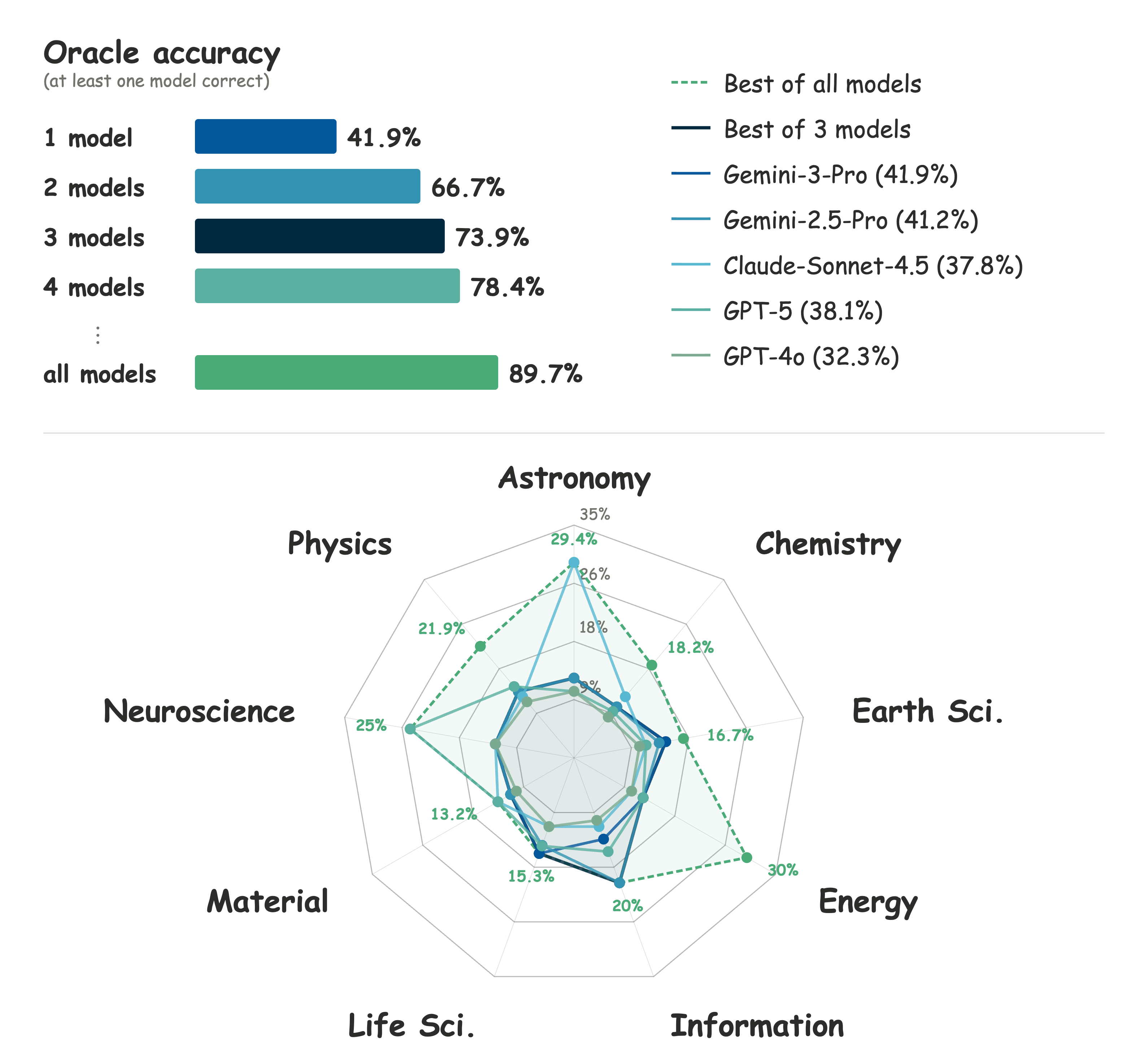}
    \caption{Frontier LLMs are complementary on the SGI-Reasoning tasks. \textit{Top}: oracle accuracy when pooling the top-$N$ frontier models. \textit{Bottom}: per-discipline accuracy across the nine fields, with the dark solid and green dashed curves marking the oracle envelopes over the best three and all evaluated models.}
    \label{fig:complementarity}
\end{figure}

\begin{figure*}[ht]
    \centering
    \includegraphics[width=\linewidth]{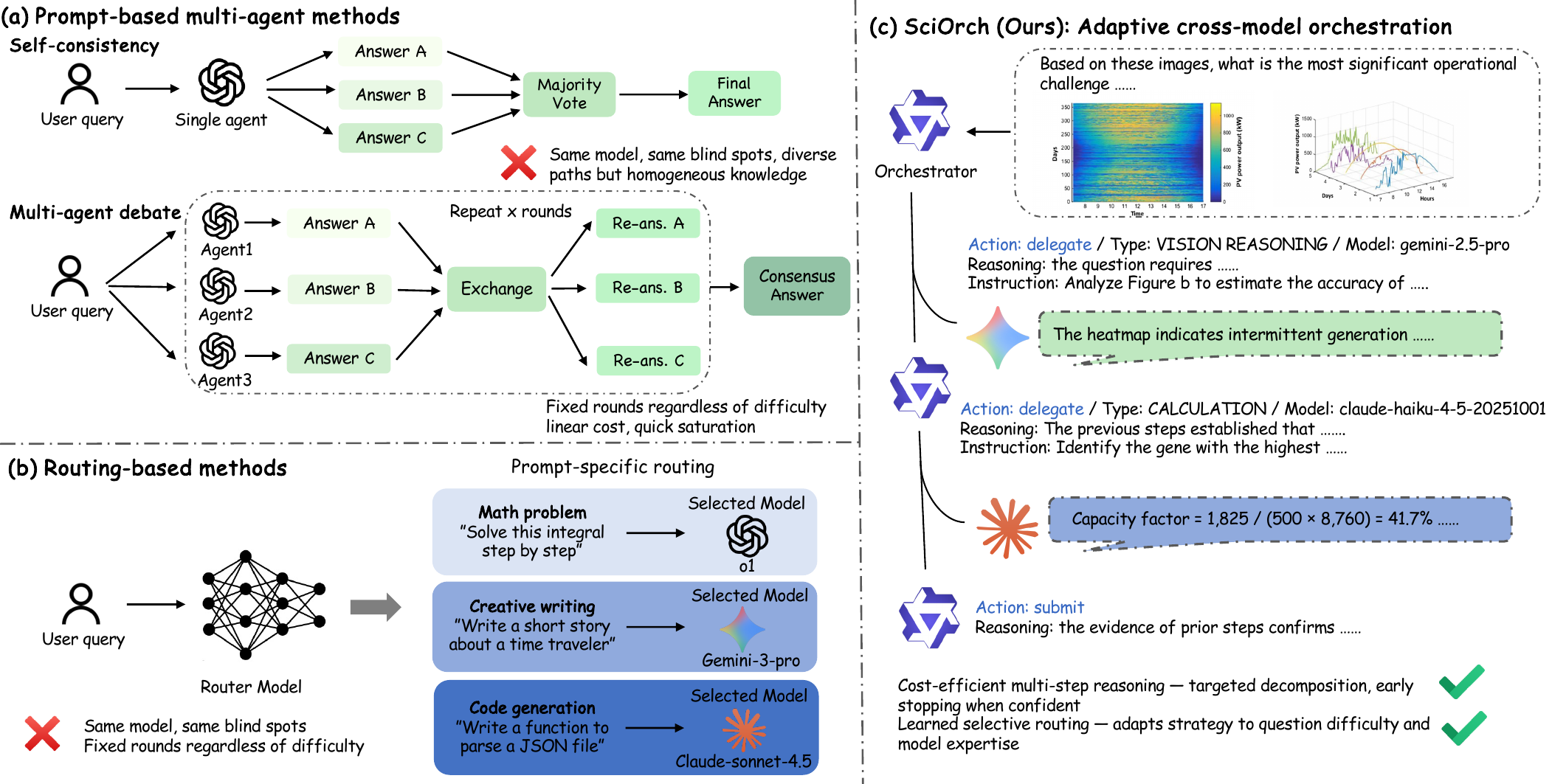}
    \caption{Comparison of multi-agent reasoning paradigms. \textbf{(a)} Prompt-based methods such as self-consistency and multi-agent debate apply a fixed protocol regardless of question difficulty, with their cost growing linearly and accuracy saturating quickly. \textbf{(b)} Routing-based methods send each query to a single pre-selected model based on prompt features, but the router can neither revise its choice after seeing intermediate evidence, nor decompose the query into sub-problems handled by different models. \textbf{(c)} \textbf{SciOrch} trains a lightweight orchestrator to decompose each question, delegate sub-problems to the most suitable commercial model, and submit a final answer once evidence is sufficient, enabling adaptive cross-model orchestration with selective routing and early stopping.}
    \label{fig:teaser}
\end{figure*}

Multimodal scientific reasoning is one of the most demanding tasks for large language models (LLMs) because of the urgent need for accelerating scientific discovery. It requires interpreting experimental figures, integrating knowledge across sub-fields, and reaching conclusions that an experienced researcher would accept. However, even the most powerful LLMs cannot reliably solve those frontier problems. SGI-Reasoning, a subset of multimodal reasoning tasks from the recent SGI-Bench~\citep{sgi-bench}, captures this setting. On the frontier tasks involving nine disciplines ranging from astronomy to materials science, even the strongest commercial model, Gemini-3-Pro~\citep{gemini3}, reaches only 42\% accuracy on this task.

Yet this ceiling masks a more nuanced picture. We evaluate a broad range of frontier models on SGI-Reasoning and find substantial disagreement among their correct answers (Figure~\ref{fig:complementarity}): combining just three models (Gemini-3-Pro, Gemini-2.5-Pro~\citep{gemini25}, and GPT-4o~\citep{gpt4o}) already covers 74\% of the test set, almost double the best single-model accuracy, and expanding the pool to all evaluated models pushes the union to 89\%. This 89\% is an oracle upper bound, since it assumes per-instance knowledge of which model is correct, but it shows how much capability already exists in the current model pool.

Different models also lead on different disciplines: Claude-Sonnet-4~\citep{claudesonnet4} ranks first on Astronomy, o4-mini~\citep{o3} on Earth Science and Physics, and Gemini-3-Pro on Life Science. The complementarity is not only across questions. A typical SGI-Reasoning question mixes figure parsing, domain-specific recall, and quantitative reasoning, and these sub-tasks do not always favor the same model: one model may read a plot more reliably while another handles the underlying mechanism better. The headroom therefore comes not from building a stronger individual model, but from combining existing ones at a granularity finer than whole-question dispatch.

Existing approaches to exploiting model diversity fall into three families: prompt-based multi-agent methods (Figure~\ref{fig:teaser}a) such as self-consistency~\citep{self-consistency} and multi-agent debate~\citep{debate}, whose protocols are static, so cost grows linearly with the number of paths while accuracy saturates regardless of question difficulty; routing-based methods (Figure~\ref{fig:teaser}b) such as prompt-to-leaderboard~\citep{p2l}, which commit each query to a single pre-selected model and can neither revise that choice after seeing intermediate evidence nor decompose the query across models; and \emph{orchestration}~\citep{toolorchestra, flowsteer, aorchestra}, where an orchestrator learns to decompose a query and route its sub-problems across models or tools. Only the last family matches the within-question complementarity above, but existing orchestration methods target general tool use and have never been extended to frontier scientific reasoning.

Inspired by orchestration-based methods, we propose \textbf{SciOrch} (Figure~\ref{fig:teaser}c), a framework that trains a lightweight 8B vision-language model (VLM) to orchestrate frontier commercial models. The orchestrator treats commercial model APIs as tools in its action space. At each step it chooses between two actions: delegate a sub-problem to a specified model, or submit a final answer. This formulation turns long-context multimodal reasoning into a multi-step decision problem. To achieve learning, reinforcement learning (RL) is the most intuitive way, but it is extremely difficult in this case for two reasons: (i) every action triggers a real API call, which makes large-scale rollouts expensive; and (ii) each call adds non-trivial latency, which makes standard online RL prohibitively slow. To achieve a balance between this performance-efficiency trade-off, we integrate Monte Carlo Tree Search (MCTS)~\citep{mcts,deepsearch} in our training. It rolls out diverse orchestration trajectories, from which we extract per-node single-turn samples and update the model with REINFORCE++~\citep{reinforce-pp}. In this way, a single rollout yields many more training samples at the same cost of API call. The gradient phase runs offline on sufficient pre-collected samples, significantly improving the training efficiency while introducing acceptable data staleness.

Empirically, SciOrch turns this complementarity into measurable gains. On the 240-question test set, it achieves the best accuracy on both evaluation slices, with 49.30\% on SGI-Reasoning~\citep{sgi-bench} and 68.10\% on Scientists' First Exam~\citep{sfe}, yielding 56.66\% average accuracy. This improves over the strongest standalone commercial model, Gemini-3-Pro, by 3.74\%, and over the strongest prompt-based multi-agent baseline by 3.33\%. The improvement is not simply purchased with more API calls: SciOrch costs only \$10.42 over the full test set, which is much lower than the best self-consistency baselines. To sum up, the core contributions of our work are three-fold:

\begin{enumerate}
    \item We propose SciOrch, an orchestration framework for frontier scientific reasoning that frames the task as a multi-step decision problem, where a lightweight agent learns to delegate sub-problems to frontier models.
    \item We introduce an MCTS-based method that makes RL tractable in the orchestration setting, where tree search generates sufficient, diverse trajectories, and offline gradient updates decouple training from API latency.
    \item Experiments on a 240-question test set across nine scientific disciplines demonstrate that SciOrch surpasses the strongest commercial model at substantially lower inference cost.
\end{enumerate}

%% file: contents/02_related_work.tex
\section{Related Work}

\subsection{Benchmarks for Scientific Reasoning}

Early benchmarks for scientific reasoning covered broad academic knowledge through multiple-choice questions and are now largely saturated by frontier models~\citep{mmlu, clark2018arc, welbl2017sciq}. Later benchmarks raised the difficulty bar with graduate-level evaluations that resist shallow retrieval, on which even the strongest reasoning models reach only 50 to 60 percent accuracy~\citep{gpqa, supergpqa, huang2024olympicarena, feng2024sciknoweval}. As vision-language models matured, a new generation of multimodal scientific benchmarks emerged to test the interpretation of figures, diagrams, and experimental data~\citep{mmmu, mmmupro, lu2022scienceqa, he2024olympiadbench, wang2024charxiv, sfe}. Among these, frontier multimodal scientific reasoning, where each question requires interpreting real experimental figures across multiple disciplines, remains the hardest, and is the setting we target through SGI-Bench.

\subsection{LLM-based Multi-agent Reasoning}

A growing body of work tries to push past the limits of any single model by combining multiple models or multiple reasoning paths~\citep{guo2024large, li2024survey, tran2025multi}. Prompt-based methods aggregate sampled reasoning paths through majority voting, debate, or structured collaboration~\citep{cot, chateval, liang2023encouraging, wang2024unleashing, qian2024scaling}; their protocols are fixed, inference cost grows linearly with the number of paths, and accuracy gains saturate quickly. A second line of work trains a router that dispatches each query to a single model from a pool, trading off answer quality against inference cost~\citep{llmrouter2025, hybridllm, frugalgpt, routerdc, graphrouter, p2l, masrouter, zhang2026patho}, but the decision is made upfront from the prompt alone and cannot adapt to intermediate evidence or decompose the query across models. A more recent direction trains a small orchestrator to delegate stronger models or tools over multiple steps~\citep{toolorchestra, flowsteer, aorchestra, zhuge2024gptswarm, zhang2024agentprune, zhang2024aflow, hu2024automated, shang2024agentsquare}, typically via RL under verifiable rewards~\citep{deepseek-r1, grpo}, sometimes combined with tree search. Most of this work targets general tool use or open-ended agentic tasks, while our setting is different, since each action triggers an expensive commercial API call rather than a cheap forward pass. We do not compare against them empirically: FlowSteer and AOrchestra target workflow design and sub-agent creation rather than expert-model orchestration, and ToolOrchestra, the closest to our setting, is text-only instead of multimodal.

\subsection{Agents for Scientific Discovery}

A separate line of research, often grouped under AI for Science, builds domain-specialized systems that act as scientific agents~\citep{ai4science2023, wang2023scientific, miret2024perspective}. Some systems train predictive models for specific scientific tasks, such as protein structure, materials, and chemistry~\citep{alphafold, alphafold3, baek2021rosetta, merchant2023scaling, szymanski2023autonomous}. Others pair LLMs with search or evolutionary procedures to discover algorithms, equations, or mathematical constructions~\citep{alphaevolve, romera2024mathematical, schmidt2009distilling}, and a third group builds full agentic pipelines that walk through hypothesis generation, experimental design, literature review, and wet-lab validation~\citep{aicoscientist, bran2023chemcrow, boiko2023autonomous, lu2024aiscientist}. These systems lean on domain-specific modules and target the full scientific process. Our work is complementary: we focus on the general reasoning step that almost any such pipeline ultimately depends on, by training a small orchestrator that combines the strengths of frontier LLMs on multimodal scientific reasoning.

%% file: contents/03_method.tex
\section{Method}

\begin{figure*}[ht]
    \centering
    \includegraphics[width=1\linewidth]{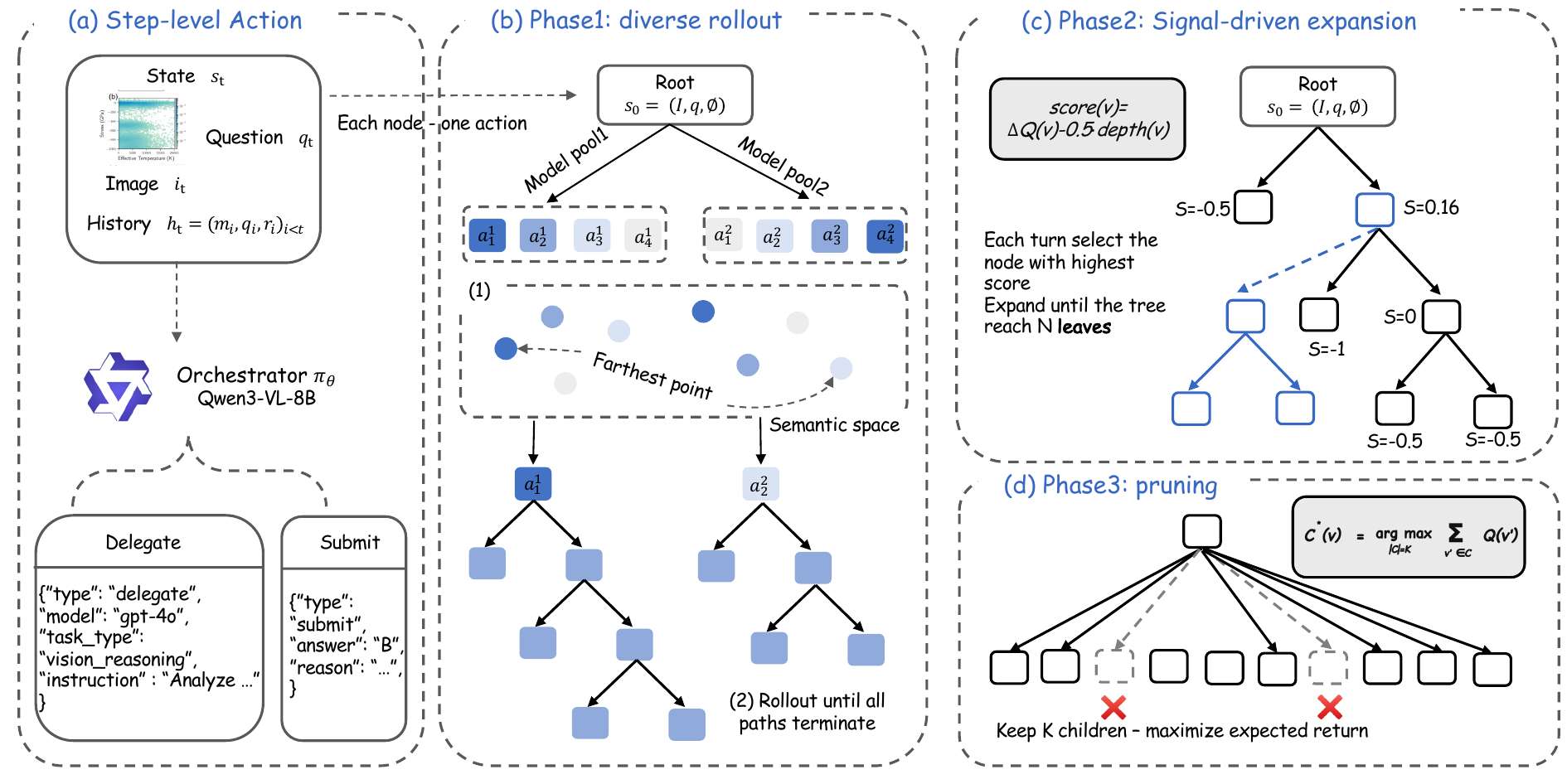}
    \caption{Three-phase MCTS for orchestrator training. \textbf{(a) Step-level action.} At each step, the orchestrator $\pi_\theta$ takes the current state $s_t = (I, q_t, h_t)$ and emits one JSON-formatted action, either a \textit{delegate} to a commercial model or a \textit{submit} of the final answer. Each node in the MCTS tree corresponds to one such action. \textbf{(b) Phase 1: diverse rollout.} Build a binary tree in two steps: (1) at each node, two prompts with different model pools each sample $n$ candidate actions, and the pair with the largest sentence-embedding cosine distance over the action text is selected as the two children; (2) rolling out from each leaf continues until all paths terminate. \textbf{(c) Phase 2: signal-driven expansion.} Grow the tree to $N$ leaves by repeatedly expanding the node with the highest $\mathrm{score}(v) = \Delta Q(v) - c \cdot \mathrm{depth}(v)$. \textbf{(d) Phase 3: pruning.} Keep at most $K$ children per parent, chosen to maximize the parent's expected return.}
    \label{fig:mcts}
\end{figure*}

\subsection{Orchestration Architecture}

We frame multimodal scientific reasoning tasks as a multi-step decision problem. Given a multimodal question $(I, q)$ with a set of images $I = \{i_1, \dots, i_n\}$ of variable size $n \geq 0$ and a text query $q$, an orchestrator $\pi_\theta$ interacts with a pool of commercial models accessed through APIs:
\begin{equation}
    \mathcal{M} = \{m_1, m_2, \dots\}.
\end{equation}
The orchestrator is a lightweight vision-language model that breaks each question into sub-problems, sends each sub-problem to a chosen model in $\mathcal{M}$, and writes a final answer from the responses it collects. Figure~\ref{fig:teaser}(c) shows the full process.

At step $t$, the state contains the original question together with what has happened so far:
\begin{equation}
    s_t = (I, q, h_t), \quad h_t = \{(m_i, q_i, r_i)\}_{i<t},
\end{equation}
where $h_t$ stores every previous delegation: which model was called, what sub-question was sent, and what response came back. Conditioned on $s_t$, the orchestrator emits a single action in a JSON-like format. The action space is:
\begin{equation}
    \mathcal{A} = \mathcal{A}_{\text{delegate}} \cup \{a_\text{submit}\}.
\end{equation}

A delegate action specifies the target model $m \in \mathcal{M}$ and an instruction $q'$ that phrases the sub-question, along with a sub-task tag drawn from \{\textsc{calculation}, \textsc{sci-reasoning}, \textsc{vision-reasoning}\}. The selected model returns a textual response $r$ via its API, and the resulting tuple $(m, q', r)$ is appended to $h_t$. A submit action carries the final answer $a$ and ends the episode. Outputs are constrained by structured prompting, and malformed actions are rejected during sampling. Figure~\ref{fig:mcts}a illustrates this step-level action structure with concrete JSON examples for both action types.

The pool $\mathcal{M}$ contains diverse frontier models with complementary strengths. The orchestrator commits to a single model per delegation instead of querying every model in parallel, so it has to learn from experience which model is good at which kind of sub-question. This is what makes the system selective routing rather than exhaustive ensembling.

\subsection{Monte Carlo Tree Search}

Training such an orchestrator with standard online RL is impractical since each rollout requires API calls. If we adopt classical online RL, the training cost and latency will explode. Therefore, we adopt MCTS to generate diverse trajectories within each iteration, so that a single tree yields many state-action samples by sharing the cost of common prefixes across branches. As shown in Figure~\ref{fig:mcts}, our MCTS runs in three phases: \emph{diverse strategy rollout}, \emph{signal-driven expansion}, and \emph{pruning}.

\noindent \textbf{Tree structure.} Multiple trajectories sampled from a task form a tree $\mathcal{G}$. Each node $v$ in the tree $\mathcal{G}$ corresponds to a state $s_v$. Edges represent actions sampled from the current policy $\pi_\theta$. The root is the initial state $s_0 = (I, q, \emptyset)$, and each leaf is either a submit node or a node that reaches the step budget.

\noindent \textbf{Phase 1: diverse rollout.} Starting from the root, we construct each node's two children in two steps (Figure~\ref{fig:mcts}b). \textbf{(1)} \textit{Candidate generation.} We issue two prompts to the orchestrator, each exposing a different subset of the model pool, so that the candidate actions cover distinct routing options at the source. Each prompt samples $n$ candidate actions, yielding $m$ candidates in total. \textbf{(2)} \textit{Farthest-pair selection.} From these $m$ candidates, we select the pair with the largest sentence-embedding cosine distance over the action text, and use them as the two children of $v$. The two steps target diversity at different stages, generation and selection, and together produce an initial tree whose siblings tend to disagree on routing choices. We continue rolling out from each leaf until all paths terminate.

\noindent \textbf{Phase 2: signal-driven expansion.} Previous works have shown that GRPO-style~\citep{grpo} methods benefit from states where different actions yield clearly different returns: the larger the return gap across siblings, the stronger the relative-advantage signal driving policy updates. We therefore introduce a signal-driven strategy that enriches such informative states until the tree contains $N$ leaves (Figure~\ref{fig:mcts}c). At each expansion step, we select the node $v$ that maximizes:
\begin{equation}
    \text{score}(v) = \Delta Q(v) - c \cdot \text{depth}(v),
\end{equation}
where $c$ is a depth-penalty coefficient and $\Delta Q(v)$ is the return gap among the children of $v$:
\begin{equation}
    \Delta Q(v) = \max_{v' \in \text{children}(v)} Q(v') - \min_{v' \in \text{children}(v)} Q(v'),
\end{equation}
and $Q(v') = \mathbb{E}_{\tau \sim v'}[R(\tau)]$ is the empirical expected return at $v'$. Parents with large $\Delta Q$ carry the strongest training signal, while the depth penalty discourages overly deep, narrow branches and keeps the tree balanced. The selected node is then expanded with the same procedure as in Phase 1.

\noindent \textbf{Phase 3: pruning.} Phase 2 may attach many children to a single parent, and some of these children can be redundant: they either share the same routing decision in different wording or fall into degenerate all-correct or all-wrong sibling groups, neither of which produces useful gradients (Figure~\ref{fig:mcts}d). For any node with more than $K$ children, we keep a subset of size $K$ chosen to maximize the parent's expected return:
\begin{equation}
    \mathcal{C}^*(v) = \arg\max_{\mathcal{C} \subseteq \text{children}(v),\ |\mathcal{C}|=K} \frac{1}{|\mathcal{C}|} \sum_{v' \in \mathcal{C}} Q(v').
\end{equation}

\subsection{Step-Level Training}

\input{tables/training_algorithm}

A straightforward use of MCTS data is to treat each root-to-leaf trajectory as one sample and apply trajectory-level policy gradients. We find this unstable in practice: trajectories sharing a common prefix receive different returns based on later decisions, so the same early action appears in samples with conflicting rewards, leading to noisy gradients.

We instead train on \emph{individual nodes} under a non-cumulative setting. For each non-leaf node $v$ in the tree $\mathcal{G}$, we form a single-turn training sample $(s_v, a_v, R_v)$, where $a_v$ is the action taken from $v$ and $R_v$ is the empirical return signal extracted from MCTS. We define $R_v$ as the value of the child reached by $a_v$, baselined against the average value of all explored siblings:

\begin{equation}
    R_v = Q(v_{\text{child}}) - V(v),
\end{equation}
where $Q(v_{\text{child}}) = \mathbb{E}_{\tau \sim v_{\text{child}}}[R(\tau)]$ is the empirical expected return at the child, and $V(v) = \frac{1}{|\text{children}(v)|} \sum_{v' \in \text{children}(v)} Q(v')$ serves as a baseline computed over siblings explored under MCTS. This decomposes the trajectory-level signal into per-step credit, with each sample reflecting how much the chosen action improves over the average alternative. Samples collected across all training questions in iteration $t$ form an on-policy training batch $\mathcal{T}_t$ that drives the policy update. The terminal return $R(\tau)$ is purely accuracy-based; an early-stage experiment with an explicit API-cost penalty yielded no meaningful savings at this training scale.

\paragraph{Policy optimization.}
Many recent RL methods for LLMs adopt GRPO and its variants, which rely on multiple rollouts from a single prompt to estimate group-relative advantages. In our setting, the relevant ``group'' is the set of sibling actions explored at the same MCTS node, and $R_v$ already encodes the within-group relative return. We therefore adopt REINFORCE++, which is naturally compatible with this per-node formulation.

For each sample $(s_v, a_v, R_v) \in \mathcal{T}_t$, we compute a token-level advantage by combining the node-level return with a per-token KL penalty against a fixed reference policy $\pi_{\text{ref}}$:
\begin{equation}
    A_v^{(\ell)} = R_v - \beta \sum_{j=\ell}^{|a_v|} \log \frac{\pi_{\theta_t}(a_v^{(j)} \mid s_v, a_v^{(<j)})}{\pi_{\text{ref}}(a_v^{(j)} \mid s_v, a_v^{(<j)})},
\end{equation}
where $a_v^{(\ell)}$ denotes the $\ell$-th token of the action, $|a_v|$ is the action length in tokens, and $\beta$ controls the strength of the KL constraint. The reference policy $\pi_{\text{ref}}$ is fixed to the initial checkpoint and prevents the orchestrator from drifting too far from a sensible language model. To stabilize updates across nodes with different return scales, we normalize advantages across the batch:
\begin{equation}
    \hat{A}_v^{(\ell)} = \frac{A_v^{(\ell)} - \text{Mean}(\{A_v^{(\ell)}\})}{\text{Std}(\{A_v^{(\ell)}\}) + \epsilon}.
\end{equation}

\noindent \textbf{Surrogate objective.} For each sample, let
\begin{equation}
    \rho_v^{(\ell)}(\theta_t) = \frac{\pi_{\theta_t}(a_v^{(\ell)} \mid s_v, a_v^{(<\ell)})}{\pi_{\theta_{\text{old}}}(a_v^{(\ell)} \mid s_v, a_v^{(<\ell)})}
\end{equation}
denote the per-token importance-sampling ratio, where $\pi_{\theta_{\text{old}}}$ is the policy used to collect $\mathcal{T}_t$. We optimize $\pi_{\theta_t}$ with a clipped surrogate objective in the style of PPO, applied at the token level:
\begin{multline}
    \mathcal{J}(\theta_t) = \mathbb{E}_{\mathcal{T}_t} \Bigg[ \sum_{\ell=1}^{|a_v|} \min\Big( \rho_v^{(\ell)}\, \hat{A}_v^{(\ell)}, \\
    \mathrm{clip}(\rho_v^{(\ell)},\, 1-\epsilon,\, 1+\epsilon)\, \hat{A}_v^{(\ell)} \Big) \Bigg].
\end{multline}
The clipping constrains each update to remain within a trusted region around $\pi_{\theta_{\text{old}}}$, which is critical for stability when training on samples that mix correct and incorrect API responses.

\noindent \textbf{Iterative training.} Each iteration $t$ consists of three phases: (1) collect an on-policy batch $\mathcal{T}_t$ by running three-phase MCTS under the current policy $\pi_{\theta_t}$ over a set of training questions; (2) compute per-token advantages $A_v^{(\ell)}$ with the KL penalty and normalize them across $\mathcal{T}_t$; (3) update $\theta$ on the clipped surrogate $\mathcal{J}(\theta_t)$. The updated checkpoint becomes the policy for the next iteration's MCTS sampling, keeping the procedure on-policy while amortizing API cost across many per-node samples. Algorithm~\ref{alg:training} summarizes the full procedure.

%% file: tables/training_algorithm.tex
\begin{algorithm}[ht]
    \caption{Training Process of SciOrch.}
    \label{alg:training}
    \begin{algorithmic}[1]
        \Require Initial policy $\pi_{\theta_0}$, training set $\mathcal{D}$, model pool $\mathcal{M}$, iterations $T$
        \Ensure Trained orchestrator $\pi_{\theta_T}$
        \For{$t = 0, 1, \dots, T-1$}
            \Statex \quad \textit{// Sampling phase}
            \State Sample a batch $\{(I_i, q_i, a_i^*)\}_{i=1}^{B} \sim \mathcal{D}$
            \State $\mathcal{T} \gets \emptyset$
            \For{each $(I_i, q_i, a_i^*)$ in the batch}
                \State $\mathcal{G}_i \gets \textsc{MCTS}(I_i, q_i, a_i^*;\, \pi_{\theta_t},\, \mathcal{M})$
                \For{each non-leaf node $v \in \mathcal{G}_i$}
                    \State $\hat{A}_v \gets Q(v_{\text{child}}) - V(v)$
                    \State $\mathcal{T} \gets \mathcal{T} \cup \{(s_v, a_v, \hat{A}_v)\}$
                \EndFor
            \EndFor
            \Statex \quad \textit{// Update phase}
            \State $\tilde{A}_v \gets (\hat{A}_v - \mu_\mathcal{T})/\sigma_\mathcal{T}$ \  \textbf{for all} $v \in \mathcal{T}$
            \State $\mathcal{L}(\theta_t) \gets -\frac{1}{|\mathcal{T}|}\sum_{v \in \mathcal{T}} \tilde{A}_v \log \pi_{\theta_t}(a_v \mid s_v)$
            \State $\theta_{t+1} \gets \theta_t - \eta \nabla_{\theta_t} \mathcal{L}(\theta_t)$
        \EndFor
    \end{algorithmic}
\end{algorithm}

%% file: contents/04_experiments.tex
\section{Experiments}

\subsection{Dataset Construction}

We construct a training and test split that combines SGI-Reasoning~\citep{sgi-bench}, our primary evaluation benchmark, with Scientists' First Exam~\citep{sfe} as a difficulty- and modality-matched augmentation source.

\noindent \textbf{SGI subset (145 train / 146 test).}
SGI-Reasoning contains 291 questions. We stratify by discipline and split into 145 training and 146 held-out test questions. The training subset preserves the discipline distribution of the full benchmark.

\noindent \textbf{SFE subset (31 train / 94 test).}
SFE contains 830 bilingual visual question-answering items. We choose SFE as the augmentation source for four reasons: (1) it is natively English with real scientific images; (2) its difficulty matches SGI-Reasoning with the o3 model~\citep{o3} reaching about 34\% accuracy on SFE, comparable to the 42\% reached by Gemini-3-Pro~\citep{gemini3} on SGI-Reasoning; (3) it shares the multimodal modality and scientific-figure framing of SGI; and (4) it does not contain training-contamination canaries.

We apply a three-step filter to extract usable samples. First, we keep only English multiple-choice items, yielding 281 candidates. Second, we apply a difficulty-matching step. Each candidate is evaluated by five frontier models, namely GPT-5.4~\citep{gpt54}, GPT-4o~\citep{gpt4o}, Gemini-3-Pro, Gemini-2.5-Pro~\citep{gemini25}, and Claude-Sonnet-4.5~\citep{claudesonnet45}, and is retained if at least one of them answers correctly. This aligns the difficulty of the augmentation source with SGI-Reasoning and keeps the training signal usable for RL, since a question no model can solve yields no reward variance within its rollout group. The filter shifts the difficulty profile of SFE, so all methods in this paper are evaluated on the same filtered subset; SGI-Reasoning itself is not filtered. Third, we sample stratified by discipline to fill the test quota first and assign the remaining items to the training set.

\noindent \textbf{Resulting split.}
Table~\ref{tab:dataset} reports the discipline composition of the final 176-question training set (145 SGI + 31 SFE) and 240-question test set (146 SGI + 94 SFE). The training set preserves the discipline weights of SGI-Reasoning, with SFE augmentation concentrated in the five disciplines it covers. The test set spans all nine SGI disciplines, and the four disciplines not covered by SFE are evaluated solely on SGI, while the others draw from both sources.

\input{tables/dataset_split}

\noindent \textbf{Baselines.} We compare SciOrch against four categories of baselines on the 240-question test set. (1) \textit{Frontier models.} We evaluate GPT-5.4, Gemini-3-Pro, and Claude-Sonnet-4.5, each prompted on the full multimodal input $(I, q)$. (2) \textit{Multi-agent methods.} We include self-consistency~\citep{self-consistency}, which majority-votes over multiple CoT samples, and multi-agent debate~\citep{debate}, where models exchange and revise answers over fixed rounds, both over the same three frontier models listed above. (3) \textit{Routing-based methods.} We compare against prompt-to-leaderboard (P2L)~\citep{p2l}, which predicts per-query Bradley-Terry win rates and selects the top model. We use the public P2L-1.5B checkpoint with its output head restricted to the same 16-endpoint pool. Since P2L is text-only, we condition it on the question stem plus OCR-extracted figure text. (4) \textit{Training-based methods.} We conduct SFT~\citep{chatgpt} and GRPO~\citep{grpo} on the Qwen3-VL-8B-Instruct model~\citep{qwen3vl} under the same training set, isolating the contribution of SciOrch's orchestration with RL-based training.

\subsection{Implementation Details}
\label{sec:implementation}

\noindent \textbf{Orchestrator and model pool.} We consistently use Qwen3-VL-8B-Instruct~\citep{qwen3vl} as the orchestrator $\pi_\theta$. In preliminary trials, a 4B variant failed to emit valid decompositions, while larger variants are impractical because MCTS sampling and RL updates scale rapidly with model size; 8B is the trade-off point we adopt. The orchestrator is paired with a model pool $\mathcal{M}$ of 16 commercial endpoints spanning three families: OpenAI (GPT-5.4, GPT-5.2, GPT-5, GPT-5-mini, GPT-4.1, GPT-4.1-mini, GPT-4o, GPT-4o-mini, o3), Anthropic (Claude-Sonnet-4.5, Claude-Sonnet-4, Claude-Haiku-4.5), and Google (Gemini-3-Pro, Gemini-3-Flash, Gemini-2.5-Pro, Gemini-2.5-Flash). The pool deliberately mixes flagship and lighter variants so that the orchestrator can also learn to trade off accuracy against cost.

\noindent \textbf{MCTS sampling.} Each Phase 1 prompt samples $n = 4$ candidate actions, so $m = 2n = 8$ per node. Phase 2 expands until the tree contains $N = 64$ leaves. Phase 3 caps each parent at $K = 8$ children. The depth-penalty coefficient is $c = 0.5$. Phase 1 diversity selection uses cosine distance over all-MiniLM-L6-v2~\citep{minilm} embeddings.

\noindent \textbf{Training.} We implement SciOrch on the basis of MS-Swift~\citep{msswift}, using its REINFORCE++~\citep{reinforce-pp} implementation as the policy optimizer. We train for $T = 12$ iterations on the 176-question training set; each iteration generates an on-policy batch via three-phase MCTS and updates the orchestrator on the clipped surrogate objective. RL starts from the base Qwen3-VL-8B-Instruct model rather than an SFT-initialized checkpoint, and that checkpoint also serves as $\pi_{\text{ref}}$. Wall-clock time is dominated by API latency, not gradient computation.

\noindent \textbf{Evaluation.} We report exact-match accuracy on three test slices, i.e., SGI, SFE, and the combined 240-question test set, along with API cost.

\subsection{Main Results}

\input{tables/main_results}

Table~\ref{tab:main-results} summarizes our findings. Vanilla Qwen3-VL-8B trails every frontier model at 27.52\% average accuracy, and training barely helps: SFT and GRPO add only 3.72\% and 0.58\%, topping out at 26.00\% on SGI and 41.30\% on SFE. 

Test-time scaling on the frontier models also fails. Self-consistency improves GPT-5.4 by only 0.42\% and debate degrades it by 2.08\%, at 5.9$\times$ and 3.2$\times$ the API spend. Claude-Sonnet-4.5 gains more, rising from 38.33\% to 46.25\% under self-consistency and 45.00\% under debate, but still trails the strongest \emph{vanilla} model, Gemini-3-Pro at 52.92\%, by 6.67 points. No single model dominates both slices: GPT-5.4 leads on SFE at 67.02\% but trails Gemini-3-Pro on SGI. The routing baseline confirms this from the other side: a preference-trained router trails the strongest frontier model by over ten points, since one upfront choice cannot exploit within-question complementarity.

SciOrch addresses both limitations. An 8B orchestrator learns to dispatch dynamically among a pool of frontier models and attains the best score on both slices at once, 49.30\% on SGI and 68.10\% on SFE. Its 56.66\% average is 3.33\% above the strongest baseline (Gemini-3-Pro with self-consistency or debate, 53.33\%) and 3.74\% above the strongest vanilla model (Gemini-3-Pro, 52.92\%). The gain is disproportionate to the cost: \$10.42 over the 240-question test set, less than half of GPT-5.4 self-consistency.

\noindent \textbf{Robustness across runs.} Over five independent runs at temperature 0.7, SciOrch averages 55.17 ($\pm$1.78)\%. One standard deviation below the mean, 53.39\%, still exceeds the strongest vanilla frontier model (52.92\%) and the strongest multi-agent baseline (53.33\%). Per-discipline cells remain too small for discipline-level significance claims.

\subsection{What Does the Orchestrator Learn?}
\label{sec:routing-analysis}

\input{tables/routing_stats}

Table~\ref{tab:routing-stats} reports, for each discipline, the expert receiving the largest share of delegations and the strongest standalone model. The routing policy is discipline-dependent rather than degenerate, and the two coincide in six of nine disciplines. The remaining three follow from sub-problem-level delegation: a model that is not the best on a whole question can still be the best at reading its figure, so the call distribution need not track whole-question accuracy. Appendix~\ref{app:routing} gives the full call distribution (Table~\ref{tab:call-distribution}) and a case study (Figure~\ref{fig:case-study}).

\subsection{Ablation Study}
\label{sec:ablation}

\input{tables/ablation}

Table~\ref{tab:ablation} ablates our three data-collection components on the first iteration, which contributes the largest single gain; a full run is too costly to repeat. \textit{Base}, the untrained orchestrator, already reaches 40.0\% by querying the expert pool.

Diverse sampling matters most: without it accuracy drops to 37.5\%, below \textit{Base}, since near-identical siblings drive $R_v$ toward zero and the update merely reinforces the prior. Removing signal-driven expansion and pruning costs 1.5 and 3.6 points respectively: the leaf budget goes to states whose siblings barely differ in return, and unpruned trees over-represent a few high-branching parents.

%% file: tables/dataset_split.tex
\begin{table}[ht]
    \centering
    \small
    \setlength{\tabcolsep}{9pt}
    \renewcommand{\arraystretch}{1.1}
    \begin{tabular}{lccc}
        \toprule
            \textbf{Discipline} & \textbf{SGI} & \textbf{SFE} & \textbf{Total} \\
        \midrule
            Astronomy    &  8/9  &  9/18 & 17/27 \\
            Chemistry    &  4/7  &  9/18 & 13/25 \\
            Earth Sci.   & 28/26 &  9/18 & 37/44 \\
            Energy       &  7/3  &   --  &  7/3  \\
            Information  &  8/12 &   --  &  8/12 \\
            Life Sci.    & 43/42 &  4/28 & 47/70 \\
            Materials Sci.     & 17/21 &  0/12 & 17/33 \\
            Neuroscience & 13/11 &   --  & 13/11 \\
            Physics      & 17/15 &   --  & 17/15 \\
        \midrule
            \textbf{Total} & 145/146 & 31/94 & 176/240 \\
        \bottomrule
    \end{tabular}
    \caption{Dataset composition by discipline and source (train/test). SFE covers five disciplines with real scientific images; ``--'' indicates the discipline is not covered by SFE, in which case both training and testing use SGI.}
    \label{tab:dataset}
\end{table}

%% file: tables/main_results.tex
\begin{table}[ht]
    \centering
    \small
    \setlength{\tabcolsep}{5pt}
    \renewcommand{\arraystretch}{1.1}
    \begin{tabular}{lcccc}
        \toprule
            \textbf{Method} & \textbf{SGI} & \textbf{SFE} & \textbf{Average} & \textbf{Cost} \\
        \midrule
            \multicolumn{5}{l}{\textit{GPT-5.4}} \\
            \quad Vanilla & 40.41 & 67.02 & 50.83 & 4.18 \\
            \quad Consistency & 45.21 & 60.64 & 51.25 & 24.73 \\
            \quad Debate & 39.73 & 62.77 & 48.75 & 13.22 \\
        \midrule
            \multicolumn{5}{l}{\textit{Claude-Sonnet-4.5}} \\
            \quad Vanilla & 31.51 & 48.94 & 38.33 & 3.14 \\
            \quad Consistency & 39.04 & 57.45 & 46.25 & 19.33 \\
            \quad Debate & 40.41 & 52.13 & 45.00 & 21.72 \\
        \midrule
            \multicolumn{5}{l}{\textit{Gemini-3-Pro}} \\
            \quad Vanilla & 46.58 & 62.77 & 52.92 & 5.97 \\
            \quad Consistency & 45.89 & 64.89 & 53.33 & 14.36 \\
            \quad Debate & 45.89 & 64.89 & 53.33 & 17.65 \\
        \midrule
            \multicolumn{5}{l}{\textit{Qwen3-VL-8B}} \\
            \quad Vanilla & 24.70 & 31.90 & 27.52 & - \\
            \quad Consistency & 22.60 & 43.62 & 30.83  & - \\
            \quad Debate & 21.23  & 44.68  & 30.42  & - \\
            \quad SFT & 26.00 & 39.40 & 31.24 & - \\
            \quad GRPO & 19.60 & 41.30 & 28.10 & - \\
        \midrule
            P2L & 32.17  & 57.50 & 42.09  & 5.39 \\
            \rowcolor{sciorchhl} \textbf{SciOrch (Ours)}   & \textbf{49.30} & \textbf{68.10} & \textbf{56.66} & \textbf{10.42} \\
        \bottomrule
    \end{tabular}
    \caption{Main results on the 240-question test set (accuracy\%). Three frontier models and Qwen3-VL-8B are evaluated under vanilla prompting, self-consistency, and multi-agent debate. We also report SFT and GRPO on Qwen3-VL-8B as training-based methods and P2L as a routing-based method. SciOrch uses Qwen3-VL-8B as an orchestrator with access to a pool of frontier models. Average is the weighted mean over 146 SGI and 94 SFE questions. SciOrch numbers are from a single run; we also report five-run statistics in the main text. Cost is the total API spend in USD.}
    \label{tab:main-results}
\end{table}

%% file: tables/routing_stats.tex
\begin{table*}[ht]
    \centering
    \small
    \setlength{\tabcolsep}{4pt}
    \renewcommand{\arraystretch}{1.1}
    \begin{tabular}{lccc}
        \toprule
            \textbf{Discipline} & \textbf{Most-called model (share)} & \textbf{Best single model} & \textbf{Match} \\
        \midrule
            Astronomy & Gemini-2.5-Pro (58.2\%) & Gemini-3-Pro & $\times$ \\
            Chemistry & Gemini-3-Pro (52.8\%) & Claude-Sonnet-4.5 & $\times$ \\
            Earth Science & Gemini-3-Pro (52.6\%) & Gemini-3-Pro / Claude-Sonnet-4.5 (tie) & \checkmark \\
            Energy & Claude-Sonnet-4.5 (100.0\%) & Claude-Sonnet-4.5 & \checkmark \\
            Information & Gemini-2.5-Pro (50.0\%) & Gemini-2.5-Pro & \checkmark \\
            Life Science & Gemini-2.5-Pro (47.6\%) & Gemini-2.5-Pro & \checkmark \\
            Materials Science & Gemini-3-Pro (51.6\%) & Claude-Sonnet-4.5 & $\times$ \\
            Neuroscience & Gemini-3-Pro (55.0\%) & Gemini-3-Pro & \checkmark \\
            Physics & GPT-4o (45.5\%) & GPT-4o & \checkmark \\
        \bottomrule
    \end{tabular}
    \caption{Per-discipline routing statistics of the trained orchestrator on the test set: the expert receiving the largest share of delegations, and the model with the highest standalone accuracy. The latter is taken within SciOrch's expert pool, a narrower scope than the full-pool ranking in Figure~\ref{fig:complementarity}. The two coincide in six of nine disciplines.}
    \label{tab:routing-stats}
\end{table*}

%% file: tables/ablation.tex
\begin{table}[ht]
    \centering
    \small
    \setlength{\tabcolsep}{7pt}
    \renewcommand{\arraystretch}{1.1}
    \begin{tabular}{lc}
        \toprule
            \textbf{Variant} & \textbf{Accuracy} \\
        \midrule
            Base (w/o training) & 40.0 \\
        \midrule
            w/o diverse sampling & 37.5 \\
            w/o MCTS expansion & 43.7 \\
            w/o pruning & 41.6 \\
        \midrule
            \rowcolor{sciorchhl} \textbf{SciOrch (1st iteration)} & \textbf{45.2} \\
        \bottomrule
    \end{tabular}
    \caption{Ablation of our MCTS data-collection procedure, run on the first training iteration (accuracy\%). \textit{w/o diverse sampling} replaces two-prompt generation and farthest-pair selection with i.i.d.\ sampling; \textit{w/o MCTS expansion} keeps the Phase 1 tree; \textit{w/o pruning} trains on all nodes without the per-parent cap.}
    \label{tab:ablation}
\end{table}

%% file: contents/05_conclusion.tex
\section{Conclusion}

In this work, we present SciOrch, a framework that trains a lightweight model to orchestrate expert models for challenging multimodal scientific reasoning. We cast frontier scientific reasoning as an \textbf{orchestration} problem and exploit the cross-model complementarity that single-model evaluation hides. To make the RL-based learning process tractable under the per-action API costs, we propose step-level MCTS-based training, which generates diverse trajectories through tree search and extracts per-node training signals for applying REINFORCE++. Experiments on a 240-question test set show that SciOrch surpasses the strongest single commercial model at lower inference cost. We hope this work motivates further study of learnable orchestration as a complementary direction to building larger individual models for frontier scientific reasoning.

%% file: contents/06_limitations.tex
\section{Limitations}

\paragraph{Evaluation scale.}
Our test set contains 240 questions across nine scientific disciplines. While this is comparable to other frontier scientific reasoning benchmarks at the same difficulty level, such as GPQA-Diamond~\citep{gpqa} with 198 questions, individual per-discipline cells are still small. We therefore report accuracy over five runs, but per-discipline trends should still be read as suggestive rather than definitive.

\paragraph{In-distribution evaluation only.}
Our test questions are held out from the same two sources used for training, so our results characterize in-distribution generalization only. To the best of our knowledge, no existing multimodal scientific reasoning benchmark matches this difficulty and figure-centric format, so a like-for-like OOD comparison is not yet available, and whether the learned routing policy transfers remains open.

%% file: contents/07_ethical_considerations.tex
\section{Ethical Considerations}

\paragraph{Potential misuse.} SciOrch lowers the cost of producing competitive scientific answers by routing a lightweight 8B orchestrator across frontier commercial APIs. While this makes scientific reasoning capability more accessible to under-resourced research groups, the same property could be exploited by bad actors. The risk is amplified by the fact that reproducing the system requires only the lightweight routing policy, not the underlying frontier weights. We urge practitioners deploying orchestration-style systems to combine them with output-side safeguards rather than relying on the cost barrier alone.

\paragraph{Model hallucination.} SciOrch's answer is synthesized from responses returned by commercial models, each of which is known to hallucinate, especially on frontier scientific questions where ground truth is scarce in pretraining data. Our orchestrator does not verify the factual content of any individual API response. It routes and aggregates rather than fact-checks. As a result, the system can confidently produce answers that are well-formatted, internally coherent, and yet factually wrong, in ways that may be harder to detect than the obvious errors a single weaker model would make. SciOrch's answer should be treated as candidate reasoning to be verified, not as  a substitute for primary literature or experimental evidence.

%% file: contents/08_appendix.tex
\input{tables/hyperparams}
\input{tables/most_call}

\section{The Use of LLMs}

We used LLMs exclusively for manuscript refinement. Specifically, we used them to correct typographical errors, fix grammatical
problems, and improve linguistic expression. The authors carefully reviewed and validated all content generated or refined with LLM assistance. We did not use LLMs for conceptual development, data collection, code implementation, experimental design, or interpretation of results.

\section{Model Documentation}

\paragraph{Orchestrator backbone.}
SciOrch uses Qwen3-VL-8B-Instruct as its policy backbone. At each decision step, the policy observes the original multimodal
question, the accumulated interaction history, and the available commercial model pool, then emits a structured action: either
delegate a sub-problem to one commercial model or submit a final answer. During training, MCTS samples candidate orchestration
trajectories using the current Qwen policy, and REINFORCE++ updates the policy from step-level signals extracted from the
resulting search tree. We also evaluate Qwen3-VL-8B-Instruct as an open-source baseline under vanilla prompting, SFT, and GRPO.

\paragraph{Expert models.}
SciOrch's expert pool $\mathcal{M}$ is the set of 16 commercial endpoints listed in Section~\ref{sec:implementation}. During training and inference, the Qwen policy delegates selected sub-problems to models in this pool through API calls, and each response is appended to the interaction history before the next decision step. Separately, GPT-5.4, Gemini-3-Pro, and Claude-Sonnet-4.5 serve as the standalone prompted baselines in Table~\ref{tab:main-results}, and self-consistency and multi-agent debate are run over these three. GPT-4o and Gemini-2.5-Pro also take part in our model-complementarity analysis and SFE quality gating, but are not reported as standalone baselines.

\paragraph{Routing baseline.}
We use the public P2L-1.5B checkpoint as the routing baseline. P2L receives the question stem together with OCR-extracted figure text, then selects one model from the same 16-endpoint pool used by SciOrch. This baseline tests whether one-shot text-based routing can match SciOrch's multi-step delegation.

\section{Dataset Documentation}

\paragraph{SGI-Reasoning.}

SGI-Reasoning is the primary benchmark used in this paper. It is a subset of SGI-Bench and contains 291 multimodal scientific reasoning questions across nine disciplines. Each example requires reasoning over scientific visual content and selecting an answer in a multiple-choice format. We stratify the dataset by discipline into 145 training questions and 146 held-out test questions. We do not apply any filtering to SGI-Reasoning because it is the target benchmark.

\paragraph{Scientists' First Exam.}

SFE is used as an augmentation source. The full dataset contains 830 bilingual visual question-answering items across Astronomy, Chemistry, Earth Science, Life Science, and Materials Science. We first keep English multiple-choice items, yielding 281 candidate examples. We then apply a difficulty-matching step with GPT-5.4, GPT-4o, Gemini-3-Pro, Gemini-2.5-Pro, and Claude-Sonnet-4.5, retaining examples that at least one frontier model answers correctly. All methods reported in this paper are evaluated on this filtered subset.

\section{Training Configuration and Cost}
\label{app:hyperparams}

Table~\ref{tab:hyperparams} lists the full configuration of SciOrch. RL training starts from the base Qwen3-VL-8B-Instruct checkpoint, which also serves as the fixed reference policy $\pi_{\text{ref}}$. 

\section{Routing Analysis}
\label{app:routing}

Table~\ref{tab:call-distribution} gives the full per-discipline call distribution. The mismatches in Astronomy, Chemistry, and Materials Science arise from sub-problem-level delegation: figure reading, quantitative reasoning, and knowledge recall are dispatched separately, so the call distribution reflects per-sub-problem suitability. Energy has too few test items for its 100\% share to be meaningful.

\section{Case Study}
\label{app:case_study}

Figure~\ref{fig:case-study} shows the full orchestration trace of SciOrch on \texttt{SGI\_Reasoning\_0033} (Life Science). Every incorrect standalone run we sampled selects the same distractor B, whose challenge clause is plausible but is not the governing difficulty. SciOrch instead splits the question: a vision delegation reads the architecture figure and rules B out on its mechanism clause, while an independent knowledge delegation, explicitly told to ignore mechanisms, rules B out on its challenge clause. Two independent lines of evidence therefore converge on C, which is the aggregate pattern of Table~\ref{tab:call-distribution} at the level of a single question.

\input{figures/case-study}

\onecolumn

\section{Prompt Templates}

\input{tables/orchestrator_prompt}
\input{tables/expert_prompt}

%% file: tables/hyperparams.tex
\begin{table}[ht]
    \centering
    \small
    \setlength{\tabcolsep}{3pt}          
    \renewcommand{\arraystretch}{1.1}
    \begin{tabular}{ll}            
        \toprule
            \textbf{Hyperparameter} & \textbf{Value} \\
        \midrule
            \multicolumn{2}{@{}l}{\textit{Orchestrator}} \\
            \quad Backbone & Qwen3-VL-8B-Instruct \\
            \quad Max steps per episode & 8 \\
            \quad Temperature (training) & 1.1 \\      
            \quad Temperature (evaluation) & 0.7 \\
        \midrule
            \multicolumn{2}{@{}l}{\textit{MCTS data collection}} \\
            \quad Candidates per prompt $n$ & 4 \\
            \quad Candidates per node $m = 2n$ & 8 \\
            \quad Leaf budget $N$ & 64 \\
            \quad Max children per parent $K$ & 8 \\
            \quad Depth penalty $c$ & 0.5 \\
            \quad Diversity embedding model & all-MiniLM-L6-v2 \\
        \midrule
            \multicolumn{2}{@{}l}{\textit{Policy optimization}} \\
            \quad Optimizer & AdamW \\
            \quad Learning rate & 1e-6 \\
            \quad KL coefficient $\beta$ & 0.01 \\
            \quad Clip range $\epsilon$ & 0.2 \\
            \quad Iterations $T$ & 12 (2 epochs $\times$ 6) \\
        \midrule
            \multicolumn{2}{@{}l}{\textit{Compute and cost}} \\
            \quad Hardware & 2 $\times$ NVIDIA H200 \\
            \quad GPU time & 26.96 GPU-hours \\
            \quad Training API spend & \$254.2 \\
            \quad Inference API spend & \$10.42 \\     
        \bottomrule
    \end{tabular}
    \caption{Full hyperparameter, compute, and cost configuration of SciOrch. Inference API spend is measured over the 240-question test set.}
    \label{tab:hyperparams}
\end{table}

%% file: tables/most_call.tex
\begin{table*}[ht]
    \centering
    \small
    \setlength{\tabcolsep}{5pt}
    \renewcommand{\arraystretch}{1.1}
    \begin{tabular}{lcccccc}
        \toprule
            \textbf{Discipline} & \textbf{Gemini-3-Pro} & \textbf{Gemini-2.5-Pro} & \textbf{GPT-4o} & \textbf{GPT-5.4} & \textbf{Claude-Sonnet-4.5} & \textbf{Others} \\
        \midrule
            Astronomy         & 10.2  & 58.2  & 8.0   & 6.0   & 9.6    & 8.0  \\
            Chemistry         & 52.8  & 8.8   & 7.2   & 5.6   & 13.6   & 12.0 \\
            Earth Science     & 52.6  & 10.0  & 7.8   & 5.8   & 11.8   & 12.0 \\
            Energy            & 0.0   & 0.0   & 0.0   & 0.0   & 100.0  & 0.0  \\
            Information       & 9.0   & 50.0  & 10.0  & 7.0   & 12.0   & 12.0 \\
            Life Science      & 8.8   & 47.6  & 9.6   & 6.8   & 12.4   & 14.8 \\
            Materials Science & 51.6  & 9.6   & 8.0   & 5.6   & 12.8   & 12.4 \\
            Neuroscience      & 55.0  & 9.0   & 8.0   & 6.0   & 11.0   & 11.0 \\
            Physics           & 9.5   & 10.0  & 45.5  & 7.0   & 13.5   & 14.5 \\
        \midrule
            Overall           & 31.5  & 27.1  & 11.6  & 6.1   & 12.3   & 11.6 \\
        \bottomrule
    \end{tabular}
    \caption{Per-discipline call distribution of the trained orchestrator on the 240-question test set. We list the five most frequently called endpoints individually; \textbf{Others} aggregates the remaining endpoints of the 16-model pool. Values are the percentage of delegate actions issued within each discipline; rows sum to 100 up to rounding. Table~\ref{tab:routing-stats} reports the strongest standalone model on each discipline for comparison.}
    \label{tab:call-distribution}
\end{table*}

%% file: figures/case-study.tex

\begin{figure*}[ht]
\centering
\small
\fbox{%
\begin{minipage}[t]{\dimexpr\textwidth-2\fboxsep-2\fboxrule\relax}
\vspace{2pt}

\textbf{Case study: \texttt{SGI\_Reasoning\_0033} (Life Science)}

\par\vspace{3pt}\hrule\vspace{4pt}
\noindent\textbf{Question.} Given tandem-MS de novo peptide sequencing data and the
RefineNovo architecture figure, identify the option that correctly states \emph{both}
the core difficulty in distinguishing peptide sequence isomers \emph{and} the mechanism
by which RefineNovo addresses it.

\smallskip
\noindent\textbf{Options.}\\
A.~``Isomers produce almost identical c/z ion distributions in ETD fragmentation;
RefineNovo captures relative intensity differences of fragment ions through a
multi-scale attention mechanism.''\\
B.~``Isomers have highly similar y-ion series in low-energy CID; RefineNovo improves
discrimination by integrating complementary b-ion information and bidirectional context
modeling.''\\
\textbf{C.~``Isomers have the same mass-to-charge ratio in MS1 precursor ion scanning;
RefineNovo simultaneously resolves all fragment ion patterns through a non-autoregressive
architecture.''}~$\star$\\
D.~``Differences in collision cross-section lead to insufficient ion mobility separation
for isomers; RefineNovo enhances feature representation by simulating collision
cross-section data.''\\
E.~``Isomers have close retention times in reverse-phase chromatography; RefineNovo
captures subtle elution profile differences through a time-series attention
mechanism.''\\
F.~``Isomers have low signal-to-noise ratios for fragment ions at low abundance;
RefineNovo enhances useful signals through a spectral denoising autoencoder.''\\
G.~``Isomers have complex charge state distributions due to multiple protonation sites;
RefineNovo improves accuracy through charge-aware fragment pattern correction.''\\
H.~``The precursor ion isotope distributions of isomers completely overlap; RefineNovo
distinguishes them through high-precision mass measurement and isotope envelope
analysis.''\\
I.~``Isomers produce similar peptide-spectrum match scores in database searches;
RefineNovo integrates multiple scoring functions through ensemble learning.''\\
J.~``Isomer identification is limited by enzymatic specificity; RefineNovo constrains
sequence search space through enzymatic cleavage probability modeling.''

\par\vspace{3pt}\hrule\vspace{4pt}
\noindent\textbf{Step 1 \quad \texttt{delegate} $\rightarrow$ Gemini-2.5-Pro \quad
\texttt{VISION\_REASONING}}

\noindent\emph{Orchestrator.} ``Inspect only the model-architecture subfigure.
What technical mechanisms are explicitly depicted for the RefineNovo-style model relative
to the vanilla NAT peptide de novo architect? State only the mechanisms visible in the
diagram [\ldots] Keep it strictly to what the figure shows.''

\noindent\emph{Expert.} ``Relative to the vanilla NAT peptide de novo architect, the
RefineNovo-style model explicitly depicts the addition of two technical mechanisms:
iterative masking and CTC path sampling.''

\medskip
\noindent\textbf{Step 2 \quad \texttt{delegate} $\rightarrow$ GPT-5.4 \quad
\texttt{SCI\_REASONING}}

\noindent\emph{Orchestrator.} ``Among the challenge descriptions embedded in
options A--J, which one is scientifically accurate for tandem mass spectrometry
distinguishing peptide sequence isomers? \emph{Ignore the RefineNovo mechanism part.}''

\noindent\emph{Expert.} ``Best match: the option whose challenge clause says peptide
sequence isomers have the same MS1 precursor $m/z$ [\ldots] Claims centered on
ETD-specific c/z behavior, retention time, CCS, isotope overlap, database scores, or
enzymatic specificity are not the core universal tandem-MS challenge here.''

\medskip
\noindent\textbf{Step 3 \quad \texttt{delegate} $\rightarrow$ Claude-Sonnet-4 \quad
\texttt{SCI\_REASONING}}

\noindent\emph{Orchestrator.} ``Using only the provided options A--J and the
evidence from the architecture figure, which option's mechanism clause is the closest
match?''

\noindent\emph{Expert.} ``Without seeing the specific options A--J, I cannot provide a
definitive letter choice. However, based on the architecture figure, the best option
would be one that describes non-autoregressive decoding enhanced by iterative masking and
CTC path sampling.''

\medskip
\noindent\textbf{Step 4 \quad \texttt{submit} $\rightarrow$ C}

\noindent\emph{Orchestrator.} ``Step 1 says the architecture figure
explicitly shows a non-autoregressive peptide decoder plus iterative masking and CTC path
sampling; this rules out option mechanisms about multi-scale attention (A), bidirectional
context (B), CCS simulation (D) [\ldots]. Step 2 says the core tandem-MS challenge is
that isomers share the same MS1 precursor $m/z$; among A--J that clause maps directly
to C. Top two considered were C and B. B captures fragment-series similarity but its
mechanism clause conflicts with the figure.''

\vspace{4pt}
\end{minipage}}
\caption{Full orchestration trace on \texttt{SGI\_Reasoning\_0033}. Each option pairs a
claimed \emph{challenge clause} (why sequence isomers are hard to separate) with a claimed
\emph{mechanism clause} (how RefineNovo addresses it); $\star$ marks the gold option C, and
B is the distractor. Text is quoted from the
logged trajectory, abridged where marked [\ldots]. The two delegations attack the distractor
from opposite sides: Step 1 reads the architecture figure and rules out B on its mechanism
clause, while Step 2 is explicitly told to ignore mechanisms and rules out B on its challenge
clause. Step 3 returned an uninformative response because the expert prompt does not forward
the answer options; the submit gate was already satisfied by Steps 1--2, and the orchestrator
proceeded without it.}
\label{fig:case-study}
\end{figure*}

%% file: tables/orchestrator_prompt.tex
\begin{tcolorbox}[title=Orchestrator Prompt, breakable, width=\textwidth]
\begin{Verbatim}[breaklines=true, breakanywhere=true, formatcom=\normalfont, fontsize=\normalsize]
You are the Orchestrator for a multimodal scientific MCQ.
You ROUTE focused sub-questions to specialist sub-models, then submit ONE final option.
Specialists answer; you decide. Do NOT solve it yourself.

Each turn, emit exactly one JSON object (schema at the end):
- delegate_task: one focused sub-question (not the whole problem).
- submit: \boxed{<letter>} (A..J).

{action_line}

Pool:
{model_guidance_text}

Routing:
- VISION_REASONING -> Vision pool. Hard CALC / SCI -> Frontier. Medium -> Strong general. Easy lookup / 1-step arithmetic -> Lightweight.
- Match by task-pool fit, not by model reputation.

Sub-question rules:
- ONE missing fact per sub-question; do not bundle multiple unknowns.
- Frame the answer so it can be mapped back to the OPTIONS.
- Do not leak your guess or prior conclusions; do not paraphrase or redefine the QUESTION.
- Do not repeat a sub-question already answered.

QUESTION:
{question}

OPTIONS:
{options}

PRIOR DELEGATE STEPS:
{prior_delegate_steps}

Reasoning (in `reasoning`, in order; keep concise):
1. Evidence: what each prior step confirmed and which OPTIONS it supports/eliminates.
2. Counter-check: ONE observation that, if found, would FLIP your current leaning. Checked?
3. Decision: if Submit Gating fails, delegate to close the gap; only then submit.

Submit Gating (ALL must hold):
(a) The decisive fact is supported by at least one delegate step (not only your own reasoning).
(b) If any prior delegate had confidence >= 0.9 on the decisive fact, either an INDEPENDENT cross-check exists (re-derive from raw values, read a different region, or apply a sanity bound), OR you justify why none is needed (e.g. direct read of a label/printed value).
(c) No unchecked counter-check could flip the answer.

Hard:
- `model` MUST be a name from the Pool above.
- `task_type` is one of CALCULATION | SCI_REASONING | VISION_REASONING.
- If a computed value is not among OPTIONS, re-examine the QUESTION intent.
- Pick the best-supported option; do not argue options are wrong.
- Only when `action` is submit: in `reasoning`, briefly name the top two letter options you considered and one sentence on why evidence favors one over the other (audit trail for the final pick only; never put this in `instruction`).
- Keep `reasoning` concise. Output must be exactly one JSON object.

Output JSON only. Exactly one of:
{"action":"delegate_task","reasoning":"...","model":"<model name>","instruction":"...","task_type":"CALCULATION|SCI_REASONING|VISION_REASONING"}
{"action":"submit","reasoning":"...","final_answer":"\\boxed{<letter>}"}
\end{Verbatim}
\end{tcolorbox}

%% file: tables/expert_prompt.tex
\begin{tcolorbox}[title=Expert Prompt, breakable, width=\textwidth]
\begin{Verbatim}[breaklines=true, breakanywhere=true, formatcom=\normalfont, fontsize=\normalsize]
You are a specialized SubAgent for multimodal scientific reasoning.

MainAgent is asking you to resolve one focused uncertainty.

Focused question from MainAgent:
{focused_question}

{task_type_hint}

Rules:
- Answer only the focused question above.
- Do not give the whole-problem final decision.
- Use the original question and images only as context.
- If the focused question requires visual evidence, examine the provided images directly and cite specific features, values, or regions you observe.
- Keep the answer concise and grounded in the task context.
- Give a local confidence for your answer to the focused question.
- Do not assume MainAgent's current direction is correct; reason independently from the evidence.
- Use a conservative confidence: low when evidence is partial/conflicting, high only when the evidence is direct and clear.
- `answer` must directly answer MainAgent's focused question.
- `confidence` must be a number in [0, 1].
- Do not output markdown fences.

Original Question:
{question}

Output JSON only:
{
  "answer": "direct answer to MainAgent's focused question",
  "evidence": "brief evidence-grounded explanation",
  "confidence": 0.00
}
\end{Verbatim}
\end{tcolorbox}